\documentclass[conference]{IEEEtran}
\IEEEoverridecommandlockouts
\usepackage{cite}
\usepackage{float}
\usepackage{amsmath,amssymb,amsfonts}
\usepackage{algorithmic}
\usepackage{graphicx}
\usepackage{textcomp}
\usepackage{xcolor}
\usepackage{booktabs}
\usepackage{multirow}
\usepackage{booktabs}

\usepackage{caption}
\usepackage{subcaption}

\def\BibTeX{{\rm B\kern-.05em{\sc i\kern-.025em b}\kern-.08em
    T\kern-.1667em\lower.7ex\hbox{E}\kern-.125emX}}
\begin{document}

%TODO LIST: check abbreviations used correctly, find fitting title, fix changes, read twice and check consistency with figure/table/abbreveations/accronyms, ~before citation.

\title{Closing the Gap in Human Behavior Analysis:\\A Pipeline for Synthesizing Trimodal Data}

\author{

 \IEEEauthorblockN{Christian Stippel, Thomas Heitzinger, Rafael Sterzinger, Martin Kampel}
 \IEEEauthorblockA{\textit{Computer Vision Lab, TU Wien}\\
 Vienna, Austria \\
 {\tt\small\{christian.stippel, thomas.heitzinger, rafael.sterzinger, martin.kampel\}@tuwien.ac.at} 
 }

}

\maketitle

\begin{abstract}
In pervasive machine learning, especially in Human Behavior Analysis (HBA), RGB has been the primary modality due to its accessibility and richness of information.
However, linked with its benefits are challenges, including sensitivity to lighting conditions and privacy concerns.
One possibility to overcome these vulnerabilities is to resort to different modalities.
For instance, thermal is particularly adept at accentuating human forms, while depth adds crucial contextual layers.
Despite their known benefits, only a few HBA-specific datasets that integrate these modalities exist.

To address this shortage, our research introduces a novel generative technique for creating trimodal, i.e., \ RGB, thermal, and depth, human-focused datasets.
This technique capitalizes on human segmentation masks derived from RGB images, combined with thermal and depth backgrounds that are sourced automatically.
With these two ingredients, we synthesize depth and thermal counterparts from existing RGB data utilizing conditional image-to-image translation.
By employing this approach, we generate trimodal data that can be leveraged to train models for settings with limited data, bad lightning conditions, or privacy-sensitive areas.
\end{abstract}


\section{Introduction}

Pervasive machine learning, particularly in Human Behavior Analysis (HBA), predominantly focuses on RGB datasets.
This modality, which is favored for its accessibility, has facilitated significant advancements in the field of HBA.
However, in some use cases, its inherent explicit visual nature is disadvantageous, e.g.,\ in settings where privacy concerns or lightning conditions play a crucial factor.

With these limitations emerges the necessity to explore other means of imagery.
Unlike RGB imagery, thermal and depth are less sensitive to lighting conditions and offer the advantage of improved highlighting of human figures.
Additionally, depth provides contextual information about the environment.
It is less susceptible to noise, making these options prime choices for sensitive HBA tasks, e.g.,\ human action recognition within privacy-critical settings.

\begin{figure}[h]
    \centering
    \begin{subfigure}[b]{0.45\textwidth}
        \centering
        \includegraphics[width=\textwidth]{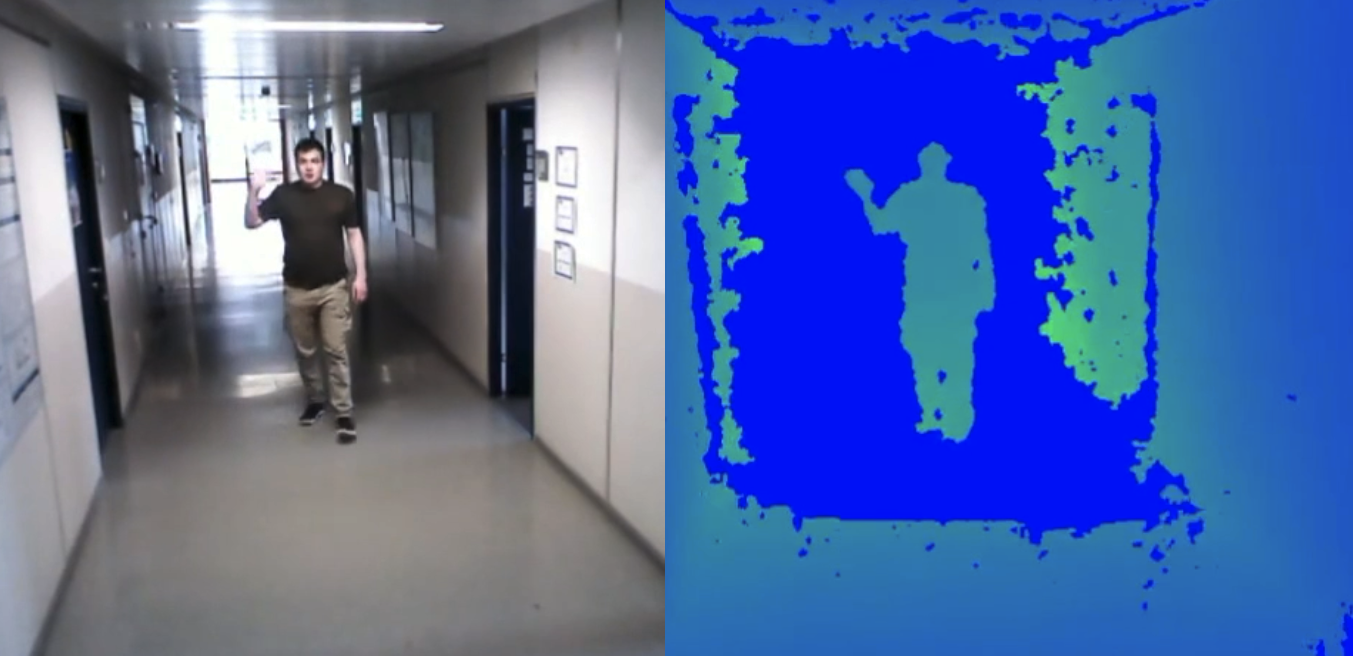}
        \caption{RGB and Depth Data in terms of privacy concerns}
        \label{fig:privacy}
    \end{subfigure}
    \hfill
    \begin{subfigure}[b]{0.45\textwidth}
        \centering
        \includegraphics[width=\textwidth]{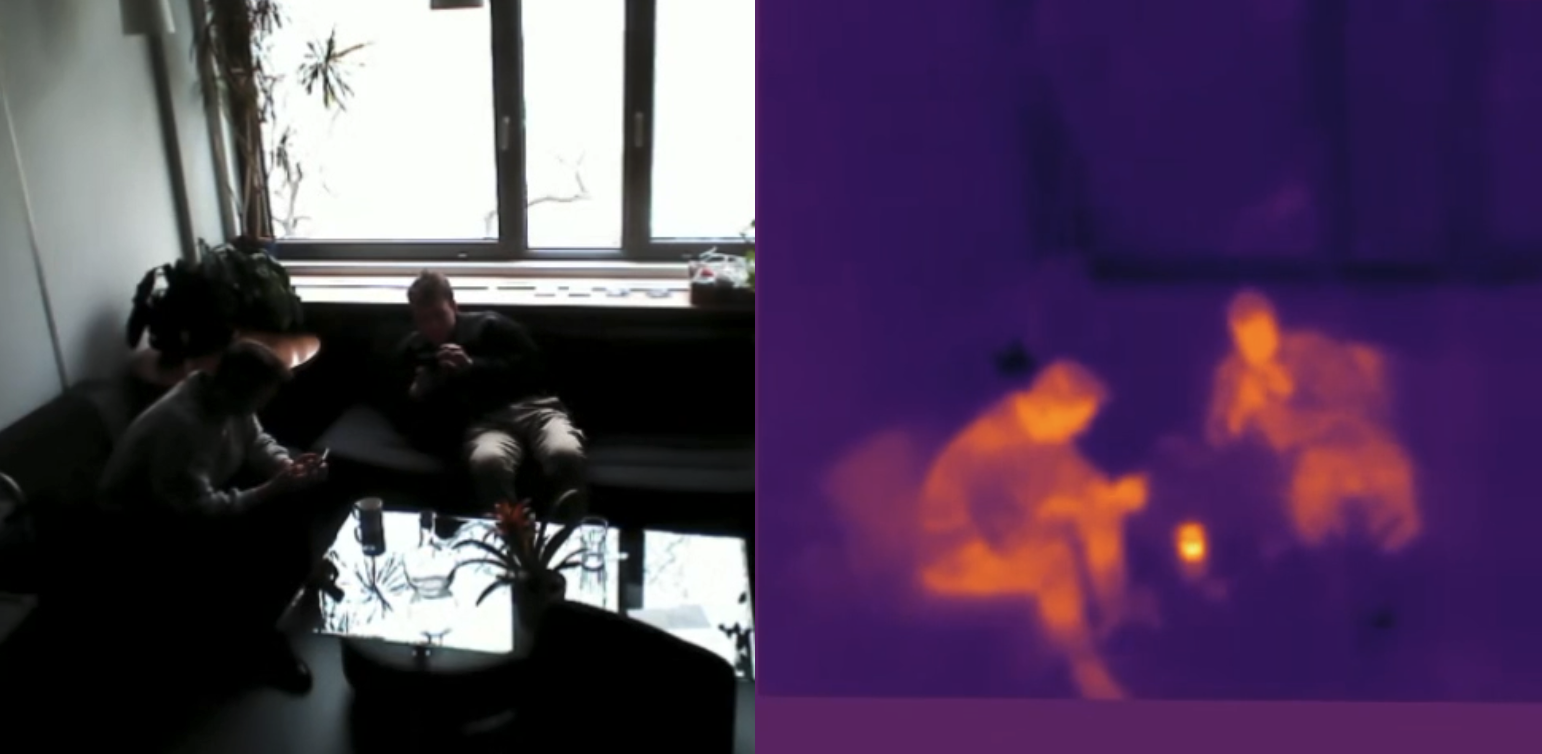}
        \caption{RGB and Thermal under difficult lighting conditions}
        \label{fig:contrast}
    \end{subfigure}
    \caption{Comparison of RGB, Thermal, and Depth Data}
    \label{fig:combined_figures}
\end{figure}

Figure \ref{fig:combined_figures} illustrates the varying levels of privacy preservation and lighting limitations offered by RGB, thermal, and depth data, highlighting the reduced privacy risks and potentially superior performance in bad lighting conditions associated with thermal and depth data.

Integrating thermal and depth data mitigates privacy and lighting sensitivity issues and paves the way for more efficient learning processes. 
These modalities can facilitate a more accurate learning of human actions by clearly outlining human forms and reducing noise \cite{stippel2023tristar, heitzinger2021bmvc}.

A notable challenge here is the relative scarcity of thermal and depth datasets compared to the abundance of RGB.
This disparity significantly hinders the potential for advanced research and applications in modalities beyond RGB.

Therefore, we present a novel image-to-image translation methodology that transforms RGB into corresponding depth and thermal data, effectively bridging the gap between these modalities.
In detail, our contributions include the following:

\begin{itemize}
    \item \textbf{Development of a Conditional Translation Pipeline}: Instead of directly mapping from RGB to depth or thermal images, our method conditions the translation process on suitable depth and thermal backgrounds.
    This approach enhances the accuracy and fidelity of the transformation since the model mainly needs to synthesize the person instead of the entire image.
    \item \textbf{Utilization of Accessible Resources}: Our pipeline leverages two upfront available resources: (1)~RGB datasets of people with corresponding action labels, captured with a static camera, and (2)~background depth and thermal frames.
    The ease of acquiring these components makes our method practical and scalable.
    \item \textbf{Dataset Generation}: By combining these resources, we efficiently generate trimodal datasets comprising RGB, depth, and thermal data to overcome the gap mentioned above effectively.
    \item \textbf{Evaluation with Action Recognition}: We evaluate our approach by training action recognition models on real and synthetic datasets.
    This demonstrates its utility in scenarios where depth and thermal data are limited.
    Additionally, it shows that our approach is an effective data augmentation strategy.
\end{itemize}

\begin{figure}[h]
  \centering
  \includegraphics[width=0.45\textwidth]{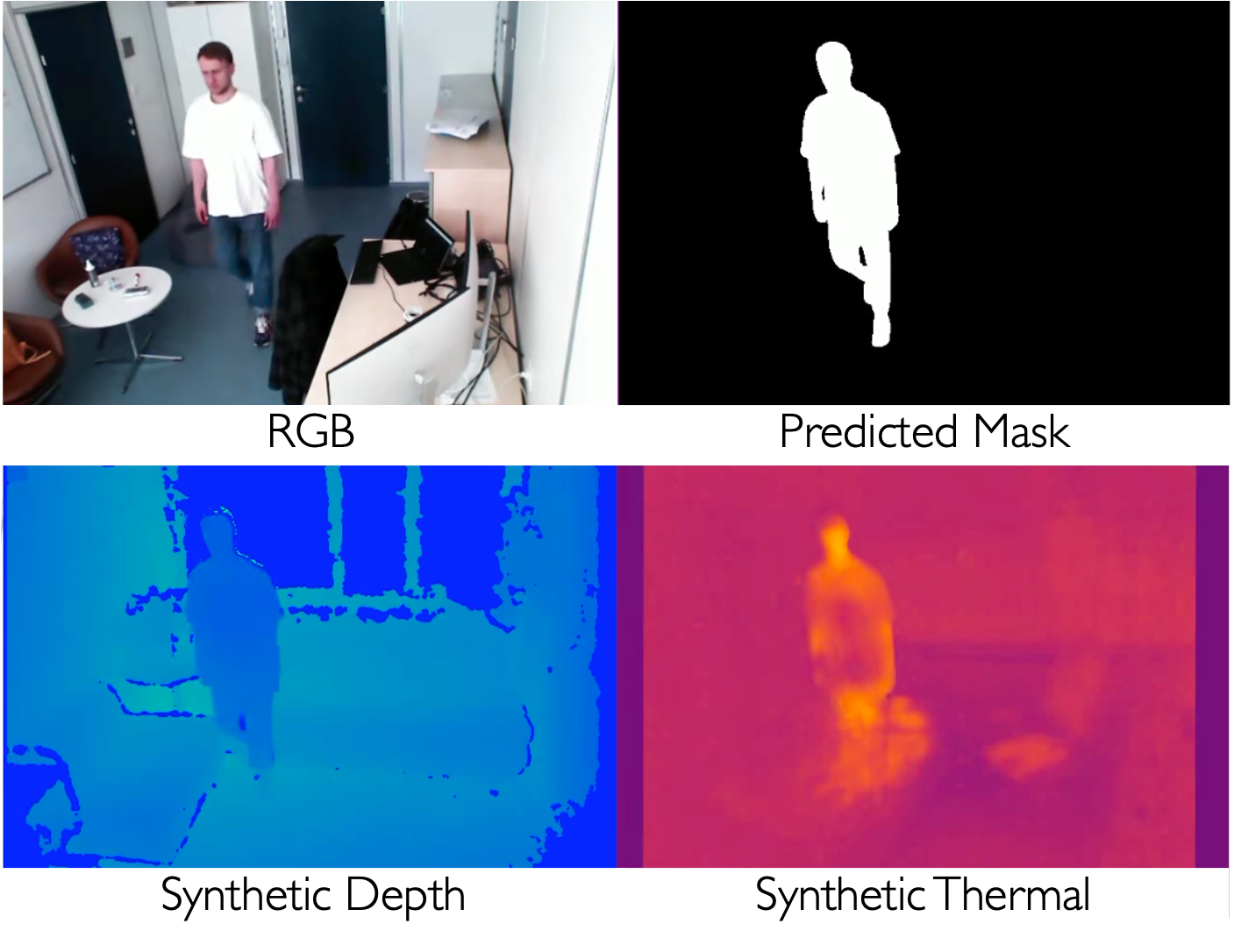}
  \caption{Illustration of the RGB to depth and thermal data transformation quality.}
  \label{fig:transformation_process}
\end{figure}

In the following, we briefly describe our proposed translation pipeline: First, we extract human segmentation masks from the given RGB data, which act as blueprints for inpainting humans onto depth/thermal background frames.
Next, we query a set of depth/thermal backgrounds to find the closest matching frame to our RGB counterpart.
Combining these two ingredients with the RGB-segmented human, we employ an image translation model to obtain the sought-after translation.
The qualitative results of this are depicted in Figure
 \ref{fig:transformation_process}, showcasing our proposed methodology's effectiveness in synthesizing high-quality depth and thermal data from RGB imagery.

Compared to prior image inpainting techniques, this process allows for the seamless integration of human figures into depth and thermal backgrounds and effectively bypasses the traditionally complex and direct conversion.

The remainder of this paper is structured as follows.
The Related Work delves into existing research on datasets across various modalities, mainly focusing on transforming data from RGB to depth and thermal modalities.
Then, we introduce and detail our translation pipeline.
This section outlines the technical framework and algorithms employed in our algorithm.
Section \ref{sec:evaluation} evaluates the effectiveness of various conditioning variations in our model.
Here, we assess how different conditioning approaches influence the model's ability to translate between modalities, offering insights into the strengths and limitations of each method.
Furthermore, we examine the potential and performance of synthetically generated datasets.
We present an analysis of synthetic data's quality, reliability, and applicability compared to real-world data, highlighting the advancements and challenges in this domain.
Finally, we conclude the paper with a summary of our findings and discuss potential avenues for future research in this field.

\section{Related Work}
\label{sec:related-work}

In general, computer vision and machine learning, the availability of high-quality datasets plays a pivotal role in advancing these fields.
For instance, datasets such as ImageNet~\cite{russakovsky2015imagenet}, PASCAL VOC~\cite{everingham2010pascal}, and COCO~\cite{lin2014microsoft} have played a vital role in pushing the limits in object detection, segmentation, and classification.
However, their focus is primarily on visual information in the RGB spectrum, which, while rich in detail, presents challenges concerning privacy and lighting conditions.

In comparison, depth datasets, such as KITTI~\cite{geiger2012we}, NYU Depth V2~\cite{silberman2012indoor}, and SUN RGB-D~\cite{song2015sun}, have contributed significantly to understanding spatial relationships in scenes.
They are particularly valuable in applications like autonomous driving, robotics, and 3D reconstruction.

Similarly, thermal datasets are crucial for applications requiring temperature mapping and night vision capabilities.
Datasets such as the OSU Thermal Pedestrian Dataset~\cite{davis2005two} offer unique insights into thermal imaging but are often limited in scope and diversity.

However, datasets consisting of more than one modality are scarce.
One example of a multimodal dataset is the KAIST Multispectral Pedestrian Dataset~\cite{hwang2015multispectral}, which combines RGB and thermal imagery for pedestrian detection.
Another more privacy-sensitive example where multimodal data plays a significant role is ``WC Buddy'' by Lumetzberger et al.~\cite{lumetzberger2021sensor}, which focuses on providing sensor-based toilet instructions for people with dementia.

This underscores a critical need in the field of computer vision: the development of comprehensive trimodal/multimodal datasets that integrate RGB, depth, and thermal to not only enrich the modalities available but also to address their limitations, e.g.,\ privacy concerns with RGB and the missing environmental context in thermal and depth.

\begin{figure*}[ht]
    \centering
    \includegraphics[width=\textwidth]{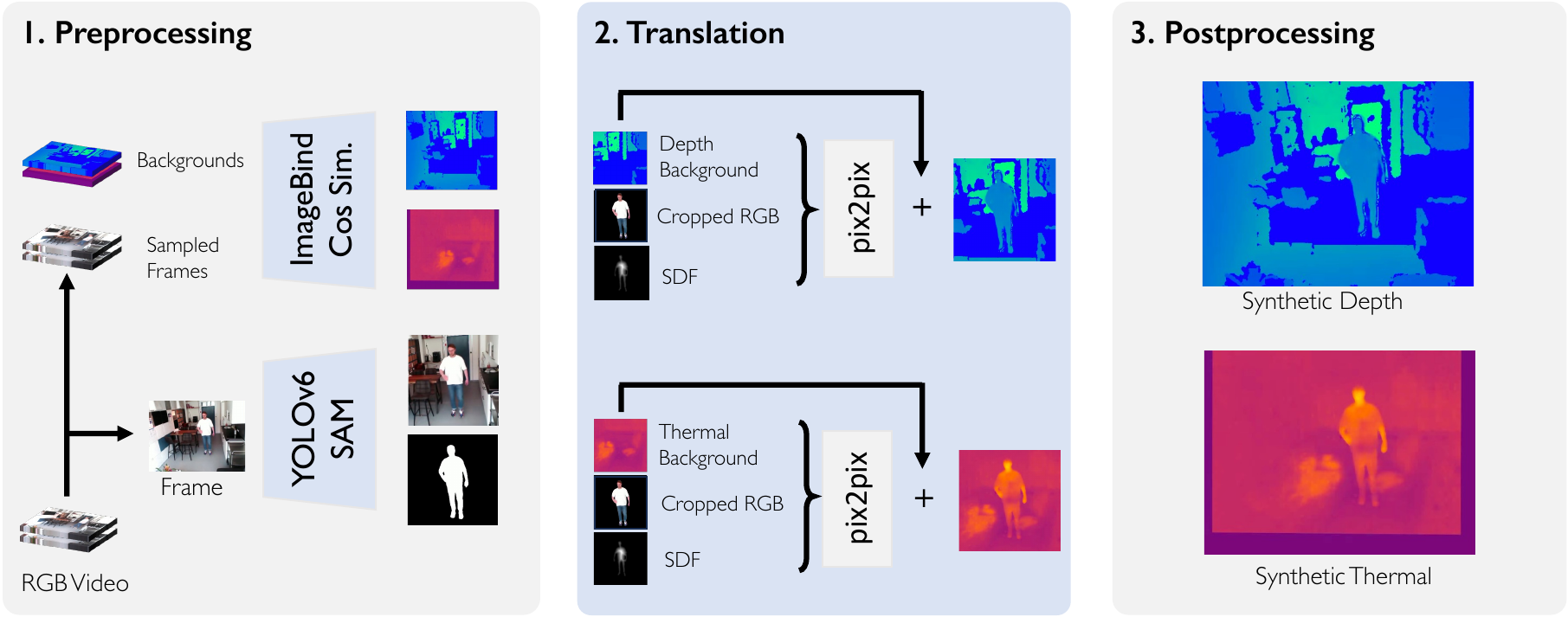}
    \caption{Overview of our proposed methodology, illustrating the integration of ImageBind for obtaining matching backgrounds, YOLOv6 and Segment Anything Model for segmenting human masks from RGB, and Pix2Pix for modality translation.}
    \label{fig:methodology_pipeline}
\end{figure*}

Image-to-image translation has emerged as a pivotal area in computer vision, focusing on converting images from one domain to another.
Pix2Pix~\cite{isola2017image} introduced a framework for paired image-to-image translation using conditional Generative Adversarial Networks (GANs).
Furthermore, specific models have been developed for translating RGB into depth and thermal:
MiDaS~\cite{ranftl2020towards}, for instance, can estimate depth from a single monocular image.
ThermalGAN~\cite{kniaz2018thermalgan}, on the other hand, focused on translating from RGB to thermal.

While there have been notable advancements in image-to-image translation models for depth and thermal, there remains room for improvement.
For instance, a notable limitation of MiDaS is that it produces inverse depth values up to a certain scale and translation.
With our proposed methodology, we argue that such unknowns can be easily inferred through contextual information given by the additional background frame.

\section{Methodology}

Our methodology transforms RGB data into depth and thermal:
First, we find matching backgrounds using ImageBind~\cite{girdhar2023imagebind} and segment human masks by combining YOLOv6 \cite{li2022yolov6} and SAM \cite{kirillov2023segment}; Second, we employ a Pix2Pix model conditioned on the background, the binary mask, and the RBG mask to create thermal and depth translations.
Figure \ref{fig:methodology_pipeline} depicts this high-level overview of our pipeline.

Our methodology contributes in the following ways:

\begin{enumerate}
    \item Simplifies the prediction task using depth and thermal backgrounds, which improves isolating the subject of interest from the environment.
    \item Employs an adjusted Pix2Pix implementation to ensure accurate translation of subject details from RGB to the target modality.
    \item Interpolates between generated details and the prepared backgrounds for seamless integration of the subject into the new modality.
    \item Conditioning on the cropped and masked RGB image. The additional RGB information gives a pixel-wise indication of object surface characteristics, such as the presence of clothing or exposed skin, which considerably narrows the distribution of plausible temperature values.
    \item Conditioning on a normalized signed distance function (SDF) to add spatial information. 
\end{enumerate}

\subsection{Locate Background}

ImageBind~\cite{girdhar2023imagebind} is a model designed to learn a joint embedding across multiple modalities, including images, text, audio, depth, thermal, and IMU.
We use this capability to locate background frames in the thermal and depth modalities that closest match the given RGB data, which we describe formally in the following:

% Defining the Embedding Functions
Let \(f_{\text{RGB}}(I)\) and \( f_{\text{thermal}}(T) \) be the functions that compute the embeddings for an RGB image~\(I\) and a thermal image~\(T\) respectively.
For a given set of RGB images \( \{I^1, I^2, \ldots, I^n\} \), the individual embeddings are given by:
\[ E_{\text{RGB}}^i = f_{\text{RGB}}(I^i)\]
Notation for a thermal embedding \( E_{\text{thermal}}^j\) follows naturally.
To obtain similarity measures for an RGB image \(I^i\) to a given set of thermal images \( \{T^1, T^2, \ldots, T^m\} \), we utilize the cosine similarity $S_C(A,B)$ and calculate a score vector as follows: 

% TODO: somehow more spacing?
\begin{align*}
\mathbf{S}^i_{\text{thermal}} = \begin{bmatrix}
S_C(E_{\text{RGB}}^i , E_{\text{thermal}}^1) \\
S_C(E_{\text{RGB}}^i , E_{\text{thermal}}^2) \\
\vdots \\
S_C(E_{\text{RGB}}^i , E_{\text{thermal}}^m) \\
\end{bmatrix}
\end{align*}

where the cosine similarity is given by:

\begin{equation}
    S_C(A, B) = \frac{A \cdot B}{\|A\| \|B\|}
\end{equation}

By aggregating scores for multiple RGB images, we obtain an average score vector of the following form:

\[
\bar{\mathbf{S}}_{\text{thermal}} = \frac{1}{n} \sum_{i=1}^{n} \mathbf{S}^i_{\text{thermal}}
\]

from which we obtain the index $\text{idx}^*_{\text{thermal}}$ of the background that is closest to our RGB images as follows:

\[\text{idx}^*_{\text{thermal}} = \arg\min(\bar{\mathbf{S}}_{\text{thermal}})\]

Similarly, a background frame can be obtained for depth or other domains.

\subsection{Obtain Human Masks}

To obtain human segmentation masks from RGB images, we employ a combination of YOLOv6~\cite{li2022yolov6}, an object detection model, and Segment Anything Model (SAM)~\cite{kirillov2023segment}, a versatile segmentation tool.
This combination can easily be adapted to other use cases, e.g., animal behavior analysis, since the underlying object detector can be swapped.

\subsection{Pre-Process Data}

\begin{figure}[h]
\centering
\includegraphics[width=0.45\textwidth]{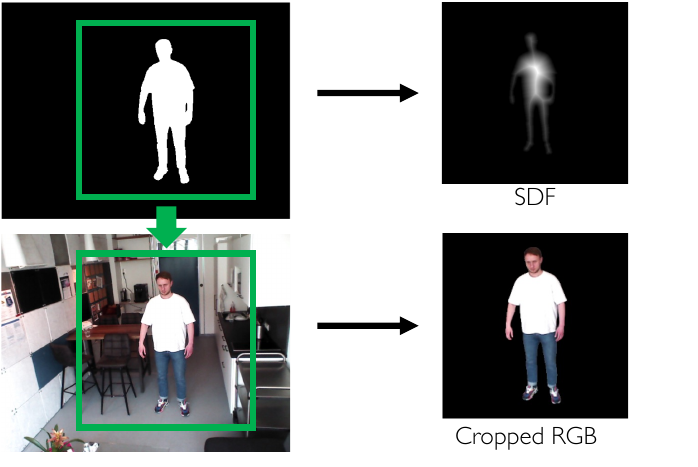}
\caption{Visualization of extracting RGB and the normalized Signed Distance Field.}
\label{fig:preprocess}
\end{figure}

Figure \ref{fig:preprocess} demonstrates the extraction and cropping of the RGB image and the normalized SDF.

First, an SDF is created based on the previously extracted human mask. 
An SDF represents a given pixel's distance to the subject's closest boundary: negative values are inside the subject, and positive values are outside.
Opposed to this, we invert it, set negative values to zero and perform min-max normalization.
We argue that this method allows our model to understand the spatial relationships within the image, i.e.\ between the person and the background, more effectively.

Next, we extract the subject from the RGB image using a padded bounding box which is based on the binary mask.
In the same manner, we crop background and SDF.
We also mask the person within the image, i.e.\ removing the background from the RGB image.
We refer to this as ``Cropped RGB'' in the translation process.
All these preprocessing steps are employed to reduce the complexity of the translation task, hence making training more stable.

Finally, we resize all inputs to a height and width of~$256$ pixels to maintain consistency across the dataset and to ensure compatibility with our translation models.

\subsection{Translate Data}

Our translation network is based on the principles of the Pix2Pix architecture~\cite{isola2017image}.
However, instead of translating within a modality, we use a five-channel input consisting of RGB, depth/thermal background, and SDF, to translate into depth/thermal.
For our backbone, we use a standard UNet architecture~\cite{ronneberger2015u}, with an EfficientNet-B4~\cite{tan2019efficientnet} serving as its encoder.
Additionally, after prediction, we manually add the background to further simplify the prediction task as the model then only needs to learn the insertion of the human figure.

During training, we optimize a joint objective: minimizing the L1- and BCE-loss.
The L1-loss is calculated based on the ground truth thermal/depth frame whereas for the BCE-loss, we integrate a PatchGAN discriminator, as described in~\cite{isola2017image}.
This discriminator evaluates local image patches which our translation model needs to fool qualitatively-wise, thereby focusing significantly more on enhancing the perceptual quality of the translated images and not only on the error L1-distance.

\subsection{Post-Process Outputs}

Lastly, our pipeline involves merging the translated cropped subject images into their corresponding backgrounds.
In detail, we merge translated images by first dilating the original mask using an \(8 \times 8\) kernel.
Then, an SDF is computed for the dilated mask, with SDF values within the original mask set to zero to focus on the border area.
This SDF is inverted and then normalized by dividing by its maximum value.
Next, masks and translated images are adjusted to align with the original image dimensions, and the translated and interpolated masks are extracted accordingly.
Finally, the translated image is merged into the original image by directly replacing pixels within the original mask and performing weighted blending at the border using the normalized SDF values to ensure a smooth transition.

In Figure \ref{fig:transformation_process}, we depict the final results of our translation methodology.

\section{Evaluation \& Results}
\label{sec:evaluation}

For evaluation, we conduct an ablation study to assess the effectiveness of conditioning our translation model on different combinations of background, masks, and RGB images as input.
Following this, we extend our evaluation to a comparative analysis, where we train an action recognition model on both synthetic and real data.
This comparison aims to demonstrate the practical applicability of our pipeline in real-world scenarios.
For this, the TRISTAR dataset by Stippel et al.~\cite{stippel2023tristar} plays a pivotal role which is utilized for multiple key aspects:

\begin{itemize}
    \item Providing background thermal and depth images for conditioning our translation models.
    \item Serving the data for training our translation models.
    \item Acting as a benchmark for our action recognition models trained on synthetic data versus those trained on real data.
\end{itemize}

\subsection{Translation Performance}
In this section, we evaluate the impact of conditioning our translation model on different combinations of inputs, namely having a background frame as input, using the normalized SDF/binary mask, using the whole/cropped RGB image, and adding the background frame after the prediction onto the performance of the model.
Here, we point out that it is to assume that performance gains based on certain input combinations will also translate well to other image-to-image translation architectures, therefore being relevant also for future advancements in the field.

The performance of the model is quantified using three key metrics: Frechet Inception Distance~(FID)~\cite{heusel2017gans}, Kernel Inception Distance~(KID)~\cite{binkowski2018demystifying}, and Mean Squared Error~(MSE).

FID and KID differ significantly from traditional metrics like MSE in several ways: They assess the quality of images based on their semantic features using a pre-trained InceptionV3 model \cite{szegedy2016rethinking}.
Instead of pixel-by-pixel comparison, they analyze the content and patterns within the images, capturing aspects such as texture, structure, and object presence.
This makes them more aligned with human perceptual evaluation.

In our study, we input normalized depth and thermal data to the Inception model.
Given that it was trained on the RGB modality, we adapt these non-RGB data by duplicating it three times to create a three-channel grayscale~``RGB'' image.
This approach allows us to use the pre-trained model without modifications.
However, since the data does not consist of standard RGB, we cut at an early layer of the model to extract lower-level features, which are more relevant and informative for our data.

In contrast to these two measures, MSE calculates the average squared difference between pixels of two images, hence lacking the ability to understand the content or context within the images.

In Table \ref{tab:depth_analysis}, we present the results of our ablation study for the depth modality.

\begin{table}[ht]
\centering
\caption{Results for Depth Analysis}
\label{tab:depth_analysis}
\begin{tabular}{@{}ccccccc@{}}
\toprule
Background & SDF & Crop & Add & FID & KID & MSE \\ \midrule
\checkmark & \checkmark & \checkmark & \checkmark & 16.20021 & 17.91997 & 0.55078 \\
\checkmark & \checkmark & \checkmark & & 34.29326 & 33.63969 & 0.53175 \\
& \checkmark & \checkmark &  & 61.38142 & 57.60419 & 1.76927 \\
\checkmark &  & \checkmark & \checkmark & 19.01336 & 20.07544 & 0.57610 \\
\checkmark &  &  & \checkmark & 22.27072 & 23.33244 & 0.57251 \\
\bottomrule
\end{tabular}
\end{table}

Combining all input conditions (adding background, SDF, and cropped RGB, as well as adding the background after the translation) results in the best performance, indicated by the lowest FID and KID scores.

This suggests that the translation under these conditions is more semantically similar to real images as lower FID and KID values signify closer alignment with the distribution of real image features.
However, the slightly worse MSE score of $\sim0.55$, while moderate, indicates that there might still be discrepancies at a per-pixel-level comparison, which is not always indicative of perceptual image quality.

On the other hand, not adding the background to the final translation slightly increases numerical accuracy but substantially increases both FID and KID.
This implies that while the model might be numerically more precise, the semantic integrity and the perceptual quality of the generated images have diminished.

Similarly, completely omitting the background while maintaining SDF and the cropped RGB leads to the highest FID, KID, and MSE scores.
This drastic increase across all metrics underscores the significance of the background in preserving both the semantic and pixel-level quality of the images.

The exclusion of specific conditions like SDF or the cropped RGB, consistently resulted in a moderate increase.
This indicates a balanced impact on both the numerical accuracy and the perceptual quality of the images, emphasizing that each condition plays a crucial role in maintaining the overall integrity and realism of the synthesized images.

In Table \ref{tab:thermal_analysis}, we depict the outcomes of our ablation study for the thermal modality.

\begin{table}[ht]
\centering
\caption{Results for Thermal Analysis}
\label{tab:thermal_analysis}
\begin{tabular}{@{}ccccccc@{}}
\toprule
Background & SDF & Crop & Add & FID & KID & MSE \\ \midrule
\checkmark & \checkmark & \checkmark & \checkmark & 1.27067 & 0.42648 & 0.35773 \\
\checkmark & \checkmark & \checkmark &  & 3.56303 & 1.78465 & 0.53721 \\
& \checkmark & \checkmark &  & 15.5289 & 9.11526 & 1.56718 \\
\checkmark &  & \checkmark & \checkmark & 1.0324 & 0.31805 & 0.53392 \\
\checkmark &  &  & \checkmark & 1.12273 & 0.34114 & 0.55877 \\
\bottomrule
\end{tabular}
\end{table}

Similar to the depth modality, having a background for the model to condition plays a crucial role in the performance and outweighs the impact of not conditioning on the cropped RGB and SDF.
This is indicated by the relatively stable FID and KID scores.
However, the numerical performance, i.e.\ the MSE when not conditioning on the cropped RGB and SDF, is worse which suggests that, while the overall semantic integrity of the images may be maintained, the precise pixel-level accuracy suffers.

\subsection{Action Recognition Performance}

We first start with a performance comparison of models trained solely on synthetic/real data.
For this we employ the 3D ConvNet proposed by \cite{stippel2023tristar} which is tailored explicitly for action recognition on the TRISTAR dataset.
Quantitatively, we assess their performance by evaluating accuracy, precision, recall, and F1 score.

The results of this are summarized in Table \ref{tab:action_recognition_performance}.
Comparing the two variants, we observe a $\sim 12\%$ reduction in F1 score when training solely on synthetic data.
Naturally, reduced performance is to be expected, and hence we regard its performance as acceptable.

\begin{table}[ht]
\centering
\caption{Action Recognition Performance (3D ConvNet)}
\label{tab:action_recognition_performance}
\begin{tabular}{ccc}
\toprule
Test Metric& Synthetic Data & Real Data \\
\midrule
Accuracy  & 0.8669 & 0.907 \\
F1   & 0.5799 & 0.707 \\
Precision & 0.6449 & 0.813 \\
Recall & 0.5268 & 0.626 \\
\bottomrule
\end{tabular}
\end{table}

Next, we integrated a second model closely aligned with the ResNext architecture~\cite{hara2017learning}.
We train this model with three different data setups: purely synthetic data, a mix of 10\% real and 90\% synthetic data, and exclusively real data.

We summarize these results in Table \ref{tab:tristar_action_recognition_performance}.
Similar to the previous experiment, training solely on synthetic data is expected to be worse.
However, the model trained with a blend of synthetic and real data achieves almost identical performance to the model trained only on real data.
This result highlights the success of our methodology as an augmentation step, especially when real trimodal data is scarce.

\begin{table}[ht]
\centering
\caption{Action Recognition Performance (ResNext)}
\label{tab:tristar_action_recognition_performance}
\begin{tabular}{cccc}
\toprule
Test Metric & Synthetic Data & Synthetic Augmentation & Real Data \\
\midrule
Accuracy  & 0.8712 & 0.90462 & 0.90409 \\ % 0.89699
F1 Score  & 0.5898 & 0.69637 & 0.69684 \\ % 0.67291
Precision & 0.6636 & 0.78246 & 0.77610 \\ % 0.75362
Recall    & 0.5307 & 0.62734 & 0.63227 \\ % 0.60781
\bottomrule
\end{tabular}
\end{table}

\section{Conclusion}

In this work, we demonstrated the effectiveness of cross-modal translation, i.e.\ from RGB to thermal and depth modalities.
Our methodology leverages a Pix2Pix model for image-to-image translation within a novel framework that adds several additions to its inputs.
The ablation study shows that adding the thermal/depth background, the normalized SDF, and cropping the RGB image, enhances the performance of the model. 
This is indicated by a reduced MSE and increased semantic quality, reflected in improved FID and KID scores.
Additionally, our action recognition experiments underscored the effectiveness of employing synthesized data within the training process: using only 10\% real data and 90\% achieves on-par performance as models trained only on real data, illustrating that our methodology can be leveraged for settings with limited data.
Lastly, the modular nature of our framework allows for the seamless integration of new components in the future.
For instance, more sophisticated alternatives, such as diffusion probabilistic models or transformer-based architectures can be employed in the future.

%\section{Acknowledgments}
%
%This work is partially funded by WWTF under the project number ICT20-
%055 for the AlgoCare project.

%%%%%%%%% REFERENCES
% \bibliographystyle{ieee_fullname}
\bibliography{references} % The filename of the .bib file without the extension
\bibliographystyle{plain} % Style of the bibliography: plain, abbrv, ieeetr, alpha, etc.

\end{document}